\documentclass[10pt,twocolumn,letterpaper]{article}

\usepackage{amsmath,amsfonts,bm}
\usepackage{gensymb}
\usepackage{mathtools}

\def\eqref#1{equation~\ref{#1}}
\def\1{\bm{1}}

\def\rmI{{\mathbf{I}}}

\def\vmu{{\bm{\mu}}}

\def\vx{{\bm{x}}}

\def\vz{{\bm{z}}}

\DeclareMathAlphabet{\mathsfit}{\encodingdefault}{\sfdefault}{m}{sl}
\SetMathAlphabet{\mathsfit}{bold}{\encodingdefault}{\sfdefault}{bx}{n}

\newcommand{\loss}{\mathcal{L}}
\newcommand{\calL}{\loss}

\newcommand{\NICKNAME}{\textsc{Gaussian3Diff}} % temp "E"ncoder "3D" "G"an inv"E"rsion
\newcommand{\nickname}{\NICKNAME} % 

\newcommand{\decoder}{D}

\newcommand{\zz}{{\bf z}}

\newcommand{\real}{\mathbb{R}}

\newcommand{\zcode}{\zz}

\newcommand{\RN}[1]{%
  \textup{\uppercase\expandafter{\romannumeral#1}}%
}

\newcommand{\bfr}{\mathbf{r}}
\newcommand{\bfx}{\mathbf{x}}

\newcommand{\bfc}{\mathbf{c}}

\newcommand{\bfmu}{\boldsymbol{\mu}}
\newcommand{\gaussian}{\mathcal{G}}
\newcommand{\payload}{{P}} % check MVP font

\newcommand{\expec}{\mathbb{E}}

\newcommand{\model}{\epsilon_\theta}
\newcommand{\denoiser}{f_\theta}

\usepackage[pagenumbers]{cvpr} % To force page numbers, e.g. for an arXiv version
\usepackage[dvipsnames]{xcolor}

\definecolor{magenta(process)}{rgb}{1.0, 0.0, 0.9}
\newcommand{\heading}[1]{\noindent\textbf{#1.}}
\usepackage{lipsum}

\newcommand\blfootnote[1]{%
  \begingroup
  \renewcommand\thefootnote{}\footnote{#1}%
  \addtocounter{footnote}{-1}%
  \endgroup
}

\definecolor{cvprblue}{rgb}{0.21,0.49,0.74}
\usepackage[pagebackref,breaklinks,colorlinks,citecolor=cvprblue]{hyperref}

\def\paperID{5192} % *** Enter the Paper ID here
\def\confName{CVPR}
\def\confYear{2024}

\title{
\nickname{}: 3D Gaussian Diffusion for 3D Full Head Synthesis and Editing
}
\author{Yushi Lan$^{1,2*}$ \quad
Feitong Tan$^{1}$ \quad
Di Qiu$^{1}$ \quad
Qiangeng Xu$^{1}$ \quad
Kyle Genova$^{1}$ \quad
Zeng Huang$^{1}$ \\
Sean Fanello$^{1}$ \quad
Rohit Pandey$^{1}$ \quad
Thomas Funkhouser$^{1}$ \quad
Chen Change Loy$^{2}$ \quad
Yinda Zhang$^{1}$ \\
$^{1}$ Google \qquad $^{2}$ S-Lab, Nanyang Technological University, Singapore
}

\begin{document}
\twocolumn[{%
\renewcommand\twocolumn[1][]{#1}%
\maketitle
    \captionsetup{type=figure}
    \vspace{-6mm}
    \centering{
    \includegraphics[width=\textwidth]{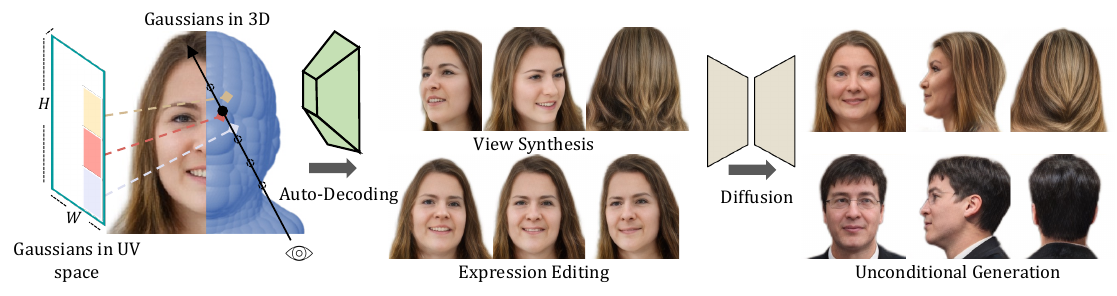}
    \captionsetup[sub]{font=large,labelfont={bf,sf}}
\hfill \vspace{-6mm}
\hfill\caption{\nickname{} adopts 3D Gaussians defined in UV space as the underlying 3D representation, which intrinsically support high-quality novel view synthesis, 3DMM-based animation and 3D diffusion for unconditional generation.}
    \label{fig:teaser}
    }
    \hfill \vspace{-0mm}
}]
% teaser
\begin{abstract}
We present a novel framework for generating photorealistic 3D human head and subsequently manipulating and reposing them with remarkable flexibility. The proposed approach leverages an implicit function representation of 3D human heads, employing 3D Gaussians anchored on a parametric face model. To enhance representational capabilities and encode spatial information, we embed a lightweight tri-plane payload within each Gaussian rather than directly storing color and opacity. Additionally, we parameterize the Gaussians in a 2D UV space via a 3DMM, enabling effective utilization of the diffusion model for 3D head avatar generation. Our method facilitates the creation of diverse and realistic 3D human heads with fine-grained editing over facial features and expressions. Extensive experiments demonstrate the effectiveness of our method.
Please check our \href{https://nirvanalan.github.io/projects/gaussian3diff/}{website}.

% Extensive experiments on various datasets demonstrate the effectiveness of our approach, resulting in notable achievements in 3D reconstruction, animation, transfer, and conditional generation.
\end{abstract}
\vspace{-2mm}    

\blfootnote{$^{*}$Work done while the author was an intern at Google.}

% \blfootnote{$^{*}$Work done while the author was an intern at Google.}
% \input{sec/1_intro}
\section{Introduction}
\label{sec:intro}

%-------------------------------------------------------------------------
% Background
Generating and editing photorealistic portraits is one of the cruxes of computer vision and graphics and has tremendous demand in downstream applications, such as embodied AI, VR/AR, digital games, and the movie industry.
Emerging neural radiance field, 3D-aware GANs~\cite{Chan2021piGANPI, Chan2021EG3D,Schwarz2020NEURIPS,gu2021stylenerf,An_2023_CVPR} have achieved great success in generating high-quality multi-view consistent portrait images with volumetric rendering.
Editing capabilities for 3D-aware GANs have also been achieved through latent space auto-decoding, altering a 2D semantic segmentation~\cite{sun2021fenerf,sun2022ide}, or modifying the underlying geometry scaffold~\cite{sun2023next3d}.
However, generation and editing quality tends to be unstable and less diversified due to the inherent limitation of GANs, and detailed-level editing is not well supported due to feature entanglement in the compact latent space or tri-plane representations.

Recently, diffusion models~\cite{song2021scorebased,NEURIPS2020_4c5bcfec} have been proposed for high-quality content generation, achieving competitive performance compared to traditional GAN-based approaches~\cite{dhariwal2021diffusion}.
Efforts have been made on 3D-aware portrait generation by de-noising on the tri-plane representation~\cite{wang2023rodin,Shue20223DNF}, which, however, do not support expression and region-based editing.

In this paper, we present \nickname{}, a diffusion-based generative model designed for 3D volumetric heads.
This model enables unconditional generation while offering versatile capabilities for both flexible global and fine-grained region-based editing, such as change of face shape, expression, or appearance.
As the core of our model, we propose a novel representation of the 3D head, in which complex volumetric geometry and appearance are encoded by a large set of 3D Gaussians modulated by tri-planes anchored on the surface of an underlying 3D face parametric model (3DMM).
We further formulate such a 3DMM surface attached representation into the UV space of the 3DMM, where each texel stores a flattened vector including 3D Gaussian parameters and the tri-plane embeddings.
We find that this representation excels, especially in geometry and expression-based editing, due to its rich semantic connection with the 3DMM model.
Furthermore,
it facilitates smooth interpolation and exchange of local or global textures, due to the dense correspondence established in the UV space.
Lastly, its 2D UV formatting ensures immediate compatibility with the well-established learning framework of 2D diffusion models~\cite{Saharia2022PhotorealisticTD}.

To this end, we propose a novel analysis-by-synthesis approach to learn a diffusion model, in which we simultaneously reconstruct large amounts of 3D heads in our representation by learning a shared latent space via an auto-decoder~\cite{park2019deepsdf} with multi-view supervision.
Compared to per-example fitting, we empirically find that the jointly optimized shared latent space encourages the alignment of local 3D Gaussians, which in turn benefits diffusion learning.
We demonstrate the effectiveness of our framework by following a DatasetGAN~\cite{zhang21datasetgan} paradigm, where the experiments are conducted on samples generated from a 3D-aware GAN, \ie, Panohead~\cite{An_2023_CVPR}, which ensures us good enough fidelity and diversity.
Trained on piles of {\it single-expression} identities only, \nickname{} achieves high-quality 3D reconstruction with the intrinsic support for 3DMM-drivable editing, and compares favorably to existing volumetric avatar generation approaches.
Furthermore, we showcase the superior editing ability of our framework with inter-subject attribute transfer, and various fine-grained editing tasks such as local region-based editing and 3D in-painting with appealing visual quality.

Our contributions are summarized as follows.
We propose a novel representation for 3D volumetric head - 3D Gaussian modulated local tri-plane on 3DMM UV space, which naively supports flexible editing capability.
We propose a novel auto-decoding-based fitting algorithm to generate training data in our representation and show it benefits diffusion model training.
Extensive experiments demonstrate that our method exhibits superior data generation quality and editing capability.

\section{Related Work}
\label{sec:related_work}
\heading{3D-aware GAN}
% ==== 2D works
Generative Adversarial Networks~\cite{Goodfellow2014GenerativeAN} have shown promising results in generating photorealistic images~\cite{karras2019style,Brock2019LargeSG,karras_analyzing_2020}
and inspired researchers to investigate using them for 3D aware generation~\cite{NguyenPhuoc2019HoloGANUL,platogan,pan_2d_2020}.
Motivated by the recent success of neural rendering~\cite{park2019deepsdf,Mescheder2019OccupancyNetwork,mildenhall2020nerf},
researchers extend NeRF~\cite{mildenhall2020nerf} to generation~\cite{Chan2021piGANPI,Schwarz2020NEURIPS,eva3d} and achieve impressive 3D-awareness synthesis.
% ==== recently
To increase the generation resolution, recent works~\cite{niemeyer2021giraffe,orel2021stylesdf,Chan2021EG3D,gu2021stylenerf,An_2023_CVPR,tan2022volux} resorted a hybrid design to high resolution up to $512$. However, samples from these methods cannot easily be edited.
On the other hand, FENeRF~\cite{sun2021fenerf} and IDE-3D~\cite{sun2022ide} proposed to generate, edit and animate human faces, guided by a segmentation map.
However, their support for local editing is still unsatisfactory, as the local geometry cannot be explicitly edited due to the lack of spatial information in the segmentation map. Additionally. Moreover, segmentation-driven animation has several limitations, \eg, can only animate an identity with similar foreface layout.
By contrast, \nickname{} achieves improved performance and flexibility via direct basic-model-driven animation.

Another line of work~\cite{Besnier2020ThisDD,Pan2022ExploitingDG,jahanian2019steerability,Jahanian2022GenerativeMA,yang2022vtoonify,zhang21datasetgan,lan2022ddf_ijcv,lan2022e3dge} propose to use a pre-trained GAN to generate training data.
Through careful design in the sampling strategy~\cite{Jahanian2022GenerativeMA}, loss functions~\cite{Pan2022ExploitingDG} and generation process~\cite{zhang21datasetgan},
off-the-shelf generators can facilitate a series of downstream applications.
In this work, we also adopt a pre-trained 3D GAN as an ``infinite'' source of 3D assets.

\heading{Diffusion Model}
Despite the remarkable success of GANs, diffusion-based models~\cite{dhariwal2021diffusion,NEURIPS2020_4c5bcfec,song2021scorebased} have recently shown impressive performance over various generation tasks, especially for 2D tasks like text-to-image synthesis~\cite{saharia2022photorealistic,rombach2022high}. 
However, applying diffusion to 3D generation is still under-explored.
Pioneering attempts have been made on shape~\cite{muller2023diffrf,chan2023genvs}, point cloud~\cite{zeng2022lion},
and text-to-3D~\cite{poole2022dreamfusion,jain2021dreamfields,dupont2022data} generation. 
Recently, some works have succeeded in training diffusion models from 2D images for unconditional 3D generation~\cite{gu2023learning,anciukevivcius2023renderdiffusion} of human faces and objects. 
However, the global 3D tri-plane in these approaches makes it difficult to edit and animate the resulting 3D representation, limiting their use for avatars.

% \yslan{
% Compare with other controllable avatar. Mention specific issues.

% Qiangeng:
% specifically, say, limitation of each previous work.

% Independent paragraph compared with existing work. 2 criteria. 
% animation performance: avatar. If fails, portrait generation.
% Also, strengthen diffusion part. At this moment, highlight 3D diffusion. Largely missing part. Propose a solution.
% }
% \input{sec/3_preliminary}
\section{Method}
\label{sec:method}

% why doing this
We propose \nickname{}, a comprehensive framework designed for the generation of photo-realistic 3D human heads with extensive editing capabilities.
% that supports expression editing. 
To fulfill this objective, we introduce a novel 3D head avatar representation in Sec.~\ref{subsec:3d-gaussian}. 
This representation leverages 3D Gaussians with local tri-planes and effectively encodes geometric and textural information in local regions. 
Critically, the 3D Gaussians are anchored to a 3D Morphable Model (3DMM), allowing for the parameterization of 3D volumetric data into the 2D texture space. This facilitates the application of a 2D diffusion model for the editing process.
In Sec.~\ref{subsec:ad-fitting}, we illustrate the diffusion-based avatar editing framework.
Initially, we delineate our analysis-by-synthesis approach that concurrently reconstructs a large number of avatars of {\it different expressions} and learns a shared latent space through multi-view supervision.
This ensures the learned representations of all avatars encapsulate crucial mutual information.
Subsequently, we account for the training of a 2D diffusion model that generates avatars with {\it neutral expression}.
In Sec.~\ref{subsec:animation-editing}, we discuss the editing mechanisms to showcase the capabilities of the proposed method.
% 
% \begin{figure*}[ht]
\begin{figure}[ht]
  \includegraphics[width=0.5\textwidth]{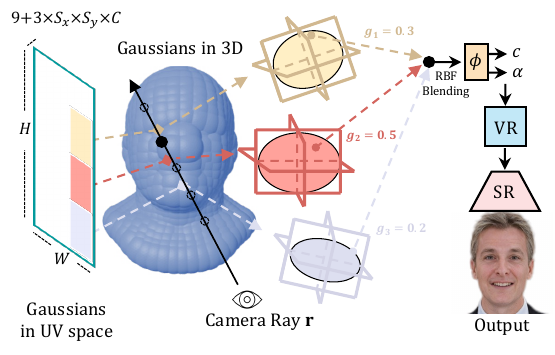} 
  \vspace{-6mm}
  \caption{During volume rendering, tri-plane payloads in UV space are projected onto 3D space with Gaussian pose parameters. For each shading point, we query the texture and geometry information from the three nearest Gaussian payloads, with influence strength defined using a radial basis function (RBF). The low-res 2D rendering is then upsampled with a CNN-based super-resolution network.}
  \vspace{-2mm}
% \end{figure*}
\end{figure}

\subsection{Avatar Representation}
\label{subsec:3d-gaussian}
\heading{3D Gaussian with tri-plane payload}
Existing methods represent a 3D head with a global representation~\cite{zheng2022imavatar,grassal2022neural,Gafni_2021_CVPR,park2021nerfies,wang2023rodin,sun2022ide}, where either a single MLP~\cite{zheng2022imavatar,grassal2022neural,Gafni_2021_CVPR,park2021nerfies} or a tri-plane~\cite{wang2023rodin,sun2022ide} is employed to encode the entire neural radiance field.
However, the global-based representation limits the region-based editing ability and cannot be directly driven by the parametric models~\cite{loper2015smpl,FLAME:SiggraphAsia2017,SMPL-X:2019}.
Inspired by previous work~\cite{Lombardi21MVP,zhang2022nerfusion,chabra2020deep} on representing radiance fields with local primitives, we propose to represent a 3D human head as a set of local tri-planes, each modulated by a 3D Gaussian initialized from a 3DMM.
%
% Specifically, the representation is composed of a group of $N$ 3D Gaussians, each of which has nine degrees of freedom, jointly optimized with a payload that encodes the geometry and texture of a small local region in 3D space.
% \begin{align}
% \label{eq:gaussian_definition}
% \gaussian = \{{\gaussian}^{i}\}_{i}^{N}, \text{where } {\gaussian}^{i} = \{ \bfmu_i, \Sigma_i,  \payload_i\},
% \end{align}
Specifically, each 3D Gaussian ${\gaussian_i}= \{ \bfmu_i, \Sigma_i,  \payload_i\}$ is characterized by 9 pose parameters and a payload - a 3D center ${\bfmu_i}$, 3 axis-aligned radii and 3 rotation angles parameterized by a 6-DOF covariance matrix $\Sigma_i$, and a tri-plane payload $\payload_i \in \real^{3\times S_x \times S_y \times C}$. These pose parameters define the local coordinate transform from the world space to the tri-plane space, as well as the influence strength. Each point $\bfx$ in the world space can be mapped to the canonical local space according to the 3D Gaussian's center ${\bfmu}$ and rotation following~\cite{zhang2023nerflets,kerbl20233d}.
The influence strength is defined as an analytic radial basis function (RBF):
\begin{equation}
\label{eq:influence_fun}
g(\bfx)=\exp \left(-\frac{1}{2}\left(\bfx-\bfmu\right)^T \Sigma^{-1}\left(\bfx-\bfmu\right)\right).
\end{equation}
% where ${\bfmu}$ is the center of the basis function and $\Sigma$ is a 6-DOF covariance matrix determined by 3 rotation angles and 3 axis-aligned radii. 

\begin{figure*}[ht!]
  \includegraphics[width=1.0\textwidth]{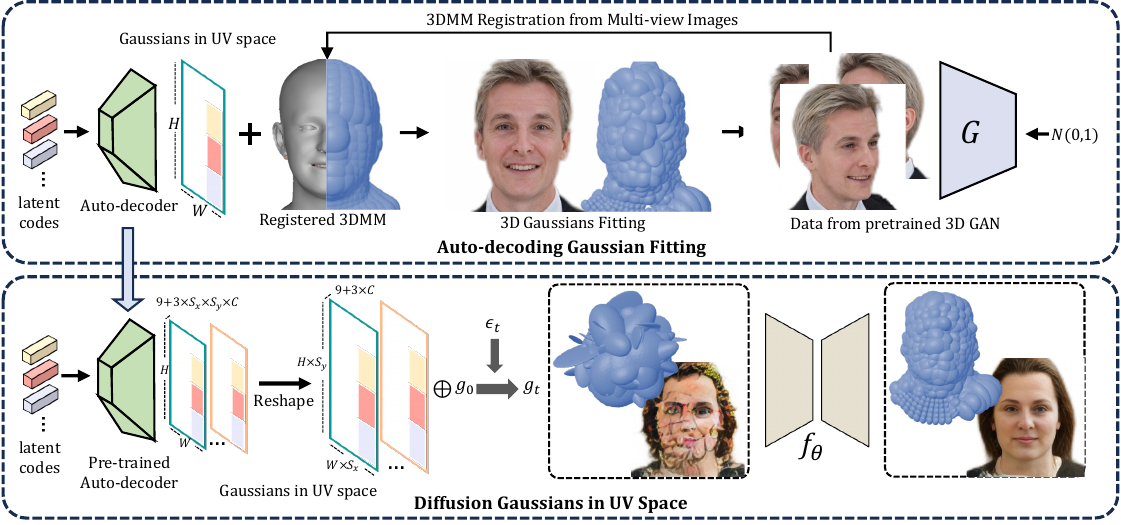} 
  \caption{Pipeline of learning 3D head generation. Top: An autoencoder is trained to generate 3D Gaussians with payloads in UV space from a dataset produced by a pretrained 3D GAN; Bottom: a diffusion model is trained by a diffusion and denoising process on the 3D Gaussians, generated by the auto-decoder trained in the previous step.}
  \vspace{-2mm}
\end{figure*}

Given a scene integrated by Gaussians, we can render any view with volumetric rendering~\cite{mildenhall2020nerf}:
\begin{align}
\label{eq:nerf_raymarching}
\hat{C}(\bfr) &=\sum_{j=1}^J T_j \alpha_j \bfc_j, 
\text{where }
 T_j =\prod_{l=1}^{j-1} (1-\alpha_l).
\end{align}
where $\hat{C}(\bfr)$ is the rendered color from the ray $\bfr$, $T_j$ is transmission at the $j$-th sample along the ray, 
$\alpha_j$, $\bfc_j$ are the opacity and color of the sample, and $J$ total number of samples along the ray.
To efficiently compute $c_j, \alpha_j$ of each sample point $\bfx_j$, we only query $K$ nearest Gaussian payloads $\gaussian_k$ measured by the Euclidean distance to the Gaussian centers ${\bfmu}_k$.
The queried features are transformed to the corresponding $\real^4$ values via a shared tiny rendering MLP $\phi$.
We then take a weighted average of the $k$ individual color and opacity by:
\vspace{-2mm}
\begin{align}
    \label{eq:nerflet_render_color}
    \bfc_j &= \sum_{k=1}^{K}\hat{g}_k(\bfx_j)\bfc_{j,k}, \\
    \label{eq:nerflet_render_alpha}
    \alpha_j &= \sum_{k=1}^{K} {g}_k(\bfx_j)\alpha_{j,k}, \\
    \label{eq:nerflet_render_g_norm}
    \text{where\ \ } \hat{g}_k(\bfx_j) &= \frac{g_k(\bfx_j)}{\sum_{k=1}^K{g_k(\bfx_k)+\epsilon}},
\end{align}
where $\bfc_{j,k}$ and $\alpha_{j,k}$ represent the color and opacity of point $\bfx_j$ queried from Gaussian $\gaussian_k$. $\hat{g}_k(\bfx_j)$ denotes the normalized inference strength and $\epsilon$ serves as a factor allowing smooth decay. 
Note that we do not normalize $g_i(\bfx_k)$ when computing opacity $\alpha_j$. This choice allows the opacity $\alpha_j$ to naturally decay in empty space. This strategy acts as a window function~\cite{lombardi2021mixture}, encouraging the Gaussians to focus on the local surface region.

In practice, we found a good balance between capacity and storage by using a total of $N=1024$ Gaussians, each associated with a local tri-plane of spatial dimensions $S_x=S_y=8$, and $C=8$ for every feature plane within the local tri-plane.
Our approach is thus more efficient than previous 3D Gaussian-based representations~\cite{kerbl20233d,keselman2022approximate} that require millions of tiny blobs where each only stores the spherical harmonic (SH) coefficients and opacity value.

\heading{UV Space Representation}
By anchoring the 3D Gaussian payloads on a 3DMM, each payload now corresponds precisely to a specific 2D location on the texture map.
Consequently, these Gaussians stored on the UV space can be processed with the U-Net-based diffusion framework~\cite{rombach2022high}. Furthermore, the semantically aligned texture map facilitates a range of editing operations.

Specifically, following previous work on the dynamic avatar reconstruction~\cite{Lombardi21MVP,bai2023learning}, we first register a 3DMM model, \eg., FLAME~\cite{FLAME:SiggraphAsia2017} for each identity instance generated from pretrained 3D GAN.  Vertices on fitted 3DMM model can be directly rasterized onto the UV space, where a 3D Gaussian is attached to each rasterized vertex.

We utilize the vertex positions to initialize $\bfmu_i$, and face normals to initialize the rotations. The axis-aligned anisotropic scaling is initialized proportionally to the area of the corresponding faces on the mesh.
Moreover, to maintain flexibility over out-of-model regions such as hair and glasses, all of the Gaussian parameters are allowed to be optimized during reconstruction.

The overall trainable parameters of each identity consist of the 9-DOF Gaussians over the UV grid: $\bfmu \in \real^{H \times W \times 3}$, $\Sigma \in \real^{H \times W \times 6}$, and the corresponding local payloads: $\payload \in \real^{H \times W \times 3\times 8 \times 8 \times 8}$.

\subsection{Learning 3D Head Generation}
\heading{Reconstruction 3D Heads with an Auto-Decoder}
\label{subsec:ad-fitting}
To effectively train the diffusion model, it is essential to have a large dataset of high-quality photorealistic 3D head assets. To address this issue, we employ the DatasetGAN~\cite{zhang21datasetgan,lan2022ddf_ijcv,lan2022e3dge} paradigm and utilize Panohead~\cite{An_2023_CVPR}, a state-of-the-art 3D GAN for generating human heads, as our data generator. This approach enables us to prepare a sufficient number of 3D assets for 3D Gaussian fitting and diffusion training.
% cannot quickly generalize to new samples

Fitting 3D assets individually involves costly reconstruction over dense multi-view images from scratch, making it data-intensive and inefficient.
% overall pipeline
To overcome this challenge, we adopt an auto-decoding design~\cite{bojanowski2017optimizing,Park_2019_CVPR,rebain2022lolnerf} that learns a shared decoder to reconstruct 3D heads by optimizing a  latent code from multi-view images.
% Technical details
Specifically, each 3D instance is associated with a latent code $\zcode \in \real^{512}$ during the optimization process. This latent code can be decoded into the local payloads in UV space through a convolutional decoder $\decoder: \real^{512} \rightarrow \real^{H \times W \times 3\times S_x \times S_y \times C}$. 
Unlike previous work~\cite{wang2023rodin} that fits tri-plane independently, our shared decoder is trained from multiple instances, enabling faster convergence and improved generalizability. Furthermore, decoding all local payloads from a shared decoder results in a smooth latent space suitable for diffusion training. 

Similar to PanoHead~\cite{An_2023_CVPR}, in order to reduce the memory consumption and computation cost, we render the color image in low resolution from 3D Ganssians and upsample them to high resolution with a super-resolution module.

During the training process, we jointly optimize all network parameters and the latent code. The loss function is decomposed into RGB loss, opacity regularization, and latent code regularization.
\begin{align}
    \label{eq:losses}
    \loss = \loss_\text{rgb} + \calL_\text{reg} + \calL_\text{code}. 
\end{align}
where ${\calL_\text{rgb}}$ is the RGB loss measured with L1 and LPIPS~\cite{zhang2018perceptual} between the synthesized color $\hat{C}$ and the ground truth color $C$ within each patch, $\calL_\text{reg}$ regularizes a compact 3D representations and ${\calL_\text{code}}$ penalize the norm of the latent code given a normal prior~\cite{bojanowski2017optimizing}.

\heading{3D Gaussian Diffusion in the UV Space}
\label{subsec:3d-diffusion}
% overall 
After the 3D Gaussians are prepared in UV space, we could learn a diffusion prior on the 3D Gaussians to support 3D avatar generation.
Specifically, a diffusion model generates data by learning the reverse of a destruction process, which is commonly achieved by gradually adding Gaussian noise over time. 
It is convenient to express the process directly in the marginals $q(\gaussian_t | \gaussian_0)$ which is given by:
\begin{equation}\small
    q(\gaussian_t | \gaussian_0) = \mathcal{N}(\gaussian_t | \alpha_t \gaussian_0, \sigma_t^2 \rmI)
\end{equation}
where $\alpha_t, \sigma_t \in (0, 1)$ are hyperparameters that determine how much signal is destroyed at a timestep $t$. Commonly, we consider variance preserving~\cite{song2021scorebased} process with $\alpha_t^2 = 1 - \sigma_t^2$. 
Before diffusion training, we drive $\gaussian_0$ to {\it neutral expression}. This ensures the subsequent generated samples can be directly manipulated using the expression basis~\cite{FLAME:SiggraphAsia2017}.

\begin{figure*}[ht]
  \includegraphics[width=1.0\textwidth]{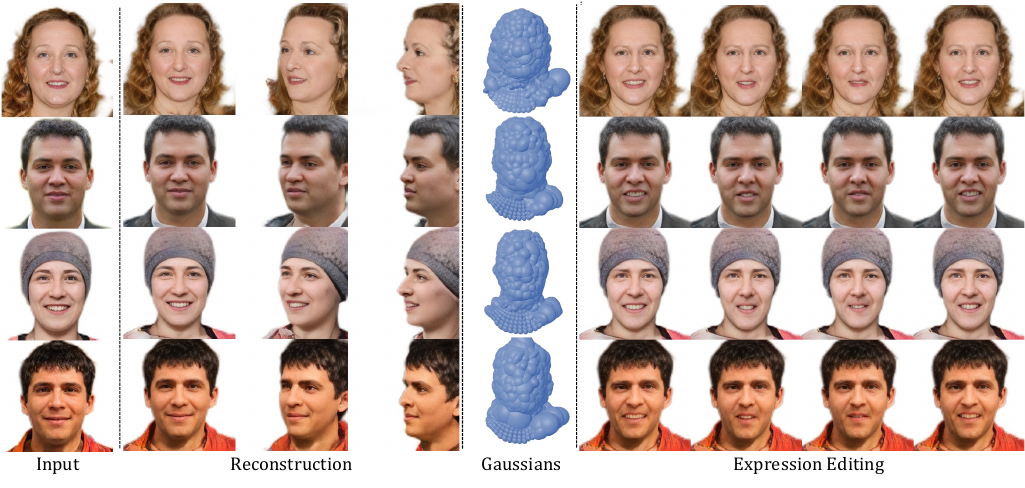} 
  \vspace{-6mm}
  \caption{\textbf{Auto-Decoder Results.} With the local Gaussians and tri-plane design, our auto-decoder $\decoder$ yield high-quality and view-consistent reconstructions. Moreover, \nickname{} intrinsically support novel expression animation by moving the positions of the optimized Gaussians. Note that we do not rely on multi-expression dataset during training.}
\label{fig:ad-results}
  \vspace{-2mm}
\end{figure*}
% diffusion equations 
\heading{Forward Process} 
Assuming the diffusion process is Markov, the forward transition is given by:
\begin{equation}\small
    q(\gaussian_t | \gaussian_s) = \mathcal{N}(\gaussian_t | \alpha_{ts} \gaussian_s, \sigma_{ts}^2 \rmI),
\end{equation}
where $\alpha_{ts} = \alpha_t / \alpha_s$ and $\sigma_{ts}^2 = \sigma_t^2 - \alpha_{t|s}^2 \sigma_s^2$ and $t > s$.
To improve the performance of the diffusion model, which favors narrower input channels~\cite{rombach2022high}, we unfold local tri-planes onto UV space along the $x$ and $y$ dimensions. This operation reshapes the Gaussian representation on UV from $\real^{W \times W \times (9+3\times S_x \times S_y \times C)}$ to $\real^{(W\times S_x) \times (W \times S_y) \times (9+3 \times C)}$,  9-d Gaussian parameters are replicated $S_x \times S_y$ times within each local tri-plane during the unfolding.

\heading{Denoising Process}
Conditioned on a single datapoint $\gaussian$, the denoising process can be written as:
\begin{equation}\small
    q(\gaussian_s | \gaussian_t, \gaussian_0) = \mathcal{N}(\gaussian_t | \vmu_{t \to s}, \sigma_{t \to s}^2 \rmI).
\end{equation}
where $\vmu_{t \to s} = \frac{\alpha_{ts} \sigma_s^2}{\sigma_t^2} \vz_t + \frac{\alpha_s \sigma_{ts}^2}{\sigma_t^2} \vx$ and $ \sigma_{t \to s} = \frac{\sigma_{ts}^2 \sigma_{s}^2}{\sigma_t^2}$. 
The literature shows~\cite{song2021scorebased} that by approximating $\gaussian_0$ by a denoiser $\hat{\gaussian_0} = f_\theta(\gaussian_t)$, we can define the learned distribution $p(\gaussian_s | \gaussian_t) = q(\gaussian_s | \gaussian_t, \vx = \hat{\gaussian_0})$ without loss of generality as $s \to t$. 

In practice, we train the denoiser $f_\theta$ to predict the input gaussian $\gaussian_0$ such that:

% ldm loss, change the L notation and prediction gt, noise schedule .etc.
\begin{equation}
\loss_\text{t}^\text{ddpm} := \expec_{\gaussian, \epsilon \sim \mathcal{N}(0, 1),  t}\Big[ w_t \Vert \gaussian_0 - f_{\theta}(\gaussian_{t},t) \Vert_{2}^{2}\Big] \, .
\label{eq:ldmloss}
\end{equation}
where the denoiser $f_{\theta}(\circ, t)$ of our model is realized as a time-conditional U-Net~\cite{ronneberger2015u}. 
We choose an empirical $w_t = \mathrm{S}(\mathrm{SNR}(t))$ where $\mathrm{SNR}(t) = \alpha_t^2 / \sigma_t^2$ and $\mathrm{S}$ is the sigmoid function, as in~\cite{gu2023learning}.
% schedule .etc.?

% \subsection{Conditioning Mechanisms}
\subsection{Editing Mechanism}
\label{subsec:animation-editing}
% \yslan{rewrite the order of this paragraph}
We emphasize three key advantages of our proposed method and explore their potential applications.
\textbf{1)}: Local Gaussians with 3DMM template.
In contrast to global-based 3D representations~\cite{park2021nerfies,park2021hypernerf,gafni2021nerface,sun2022ide,gu2023learning,An_2023_CVPR} where each attribute is intricately entangled, our method gains advantage by integrating 3D scenes with local Gaussians.
This approach allows for isolating and controlling local edits without unintended propagation to the global representation.
Additionally, anchoring the Gaussians over 3DMM inherits the benefits of a 3DMM, enables direct identity and expression editing. 
\textbf{2)}: UV-space parameterization. 
By rasterizing the 3DMM onto semantically consistent UV space, our method facilitates flexible region-based editing
Specifically, we can directly transfer~\cite{lan2022ddf_ijcv} specific semantic regions, such as the mouth or nose, across identities by swapping their learned local Gaussians. Leveraging the trained diffusion model, we can further edit the region by diffusing the masked region in UV space to while keeping the remaining areas frozen. 
\textbf{3)} Geometry-texture disentanglement.
Empowered by floatable 3D Gaussians with tri-plane payload, a noteworthy byproduct benefits of \nickname{} is the support of geometry-texture disentanglement.
All the aforementioned editing applications can be conducted on either geometry, texture, or both.

\section{Experiment}

\label{sec:experiment}

\heading{Dataset}
To maintain both quality and diversity, we sample $10,000$ 3D portraits from pre-trained Panohead~\cite{An_2023_CVPR} with diverse identities and expressions.
For each identity, we render $50$ multi-view images and depths with known camera poses. 
We use the $64 \times 64$ view-consistent 3D renderings for Gaussian fitting, and the $512 \times 512$ samples for super-resolution training.
% 
% All methods are retrained from scratch on this dataset for fair comparisons.
% For each sampled 3D identity, 
We filter out the low-quality samples 
% with Janus problem~\cite{poole2022dreamfusion} 
using CLIP~\cite{Radford2021CLIP}.
        
\heading{Implementation Details}
We use $N=1024$ Gaussians to represent each 3D identity, given $H=W=32$. During rendering, we adopt $K=3$ for nearby Gaussian blending. 
The autodecoder $\decoder$ is implemented similarly to StyleGAN~\cite{karras2019style} with noise injection removed.
After the $\decoder$ is trained, we further stack a $\times 4$ super-resolution model above it with the architecture from ESRGAN~\cite{wang2018esrgan}.
The denoiser $\denoiser$ is implemented as a 2D U-Net with architecture from Imagen~\cite{Saharia2022PhotorealisticTD}. 
The decoded UV maps of all instances are exported from the trained autodecoder $\decoder$ as the training corpus of the diffusion model $\denoiser$.
We use $2$ A6000 GPUs for model training.
% 
% Please check the supplementary material for more details.

\heading{Evaluation Metrics}
We select a series of proxy metrics to benchmark our method.
Following~\cite{sun2022ide}, we evaluate view consistency assessed by multi-view facial identity consistency (ID)~\cite{deng2019arcface} rendered from random camera poses.
To evaluate the synthesized 3D geometry, we follow EG3D~\cite{Chan2021EG3D} to use an off-the-shelf tool to estimate depth maps from renderings and compute L2 distance against rendered depths.
Moreover, we adopt an avatar-centric metric, Percentage of Correct Keypoints (PCK)~\cite{Yang2013ArticulatedHD} to evaluate the expression editing ability.
The rendering speed and storage are also included.
\begin{table}[tp]
\centering
\small
{
\caption{\textbf{Quantitative performance.} \nickname{} achieves competitive performance over 3D-related metrics (ID, Depth) and SoTA performance on the expression editing (PCK) performance. Additionally, yields faster rendering with less storage required with competitive 3D metrics.} 
\vspace{-2mm}
\label{tab:quant}
\resizebox{\linewidth}{!}{
\begin{tabular}{l@{\hspace{3mm}}ccccccc}
\toprule
Methods & ID  $\uparrow$ &  Depth $\downarrow$ &   PCK@2.5 $\uparrow$  & PCK@5$\uparrow$   & FPS $\uparrow$ & Storage(MB)$\downarrow$  \\
\midrule
FENeRF & 0.61 & 2.71 & - & - & 1.2 & 10 \\
Panohead & \textbf{0.80} & {2.32} & - & - & 19  & 72 \\
IDE-3D & 0.76 & \textbf{1.71} & 0.16 & 0.33 & 25.1 & 48 \\
\midrule
Ours    &{0.78} & 2.58 & \textbf{0.783} & \textbf{0.99} & \textbf{47/27} & \textbf{8.25} \\
 \bottomrule
\end{tabular}}}
\vspace{-2mm}
\end{table}

\subsection{Quantitative Comparisons}
\label{subsec:quantitative-eval}
The results of numerical comparisons are presented in Tab.~\ref{tab:quant}. 
Given that our method leverages Panohead data for training, it exhibits similar performance on ID and Depth metrics.
In terms of expression editing ability, 
conventional global-based methods such as FeNeRF and Panohead do not support animation.
Though IDE-3D supports segmentation-based reenactment, it lacks identity preservation and falls behind the PCK metric.
\nickname{} stands out as the only method that supports 3DMM-driven expression editing and achieving better PCK performance under both thresholds.
Moreover, \nickname{} supports faster rendering (47 FPS w/o SR and 27 FPS with SR) with less storage required for synthesized human heads.

\begin{figure*}[ht]
  \includegraphics[width=1.0\textwidth]{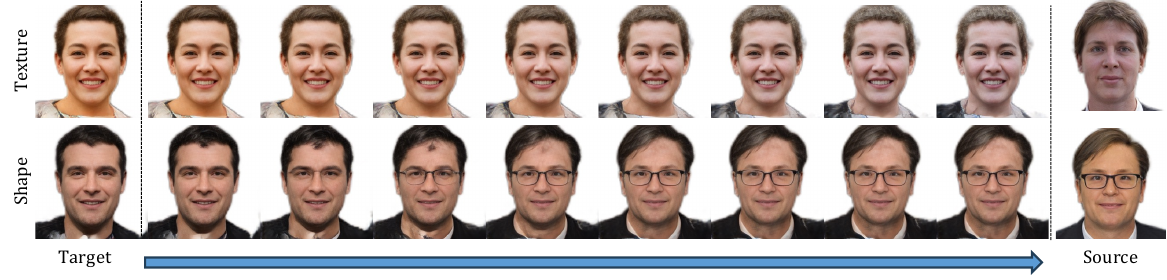}
\vspace{-6mm}
  \caption{\textbf{Shape-Texture Interpolation.} 
  We visualize the intermediate trajectory of texture transfer in row$-1$, and the shape transfer in row$-2$. Both shape and texture interpolation results preserve high-fidelity during the middle state.
  }
\label{fig:geo-tex-interp-trajectory}
\vspace{-2mm}
\end{figure*}
\subsection{Qualitative Evaluations}
\label{subsec:qualitative-eval}
% \heading{Gaussian Fitting}
% \yslan{visualization of the learned Gaussian.}
\heading{Auto-decoded Gaussians}
We first visualize the Gaussians reconstruction from the autodecoder $\decoder$ in columns $1-4$ of Fig.~\ref{fig:ad-results}.
The reconstruction produces high-fidelity and view-consistent view synthesis.
Additionally, the corresponding optimized Gaussians align with the identity shape of the input, showcasing the robust capacity of our design.

\heading{Expression Editing}
We further include the novel expression editing performance in columns $5-8$ Fig.~\ref{fig:ad-results}. Despite being trained on collections of identities with a single expression, \nickname{} inherently supports 3DMM-based expression editing by manipulating the underlying 3D Gaussians.
Furthermore, owing to the autodecoder design, \nickname{} can learn diverse expressions across identities, yielding natural-looking results under novel expressions.

\heading{Shape-Texture Transfer}
\nickname{} naturally supports geometry-texture disentanglement, where Gaussians managing the geometry and attached local tri-planes determining the texture within a local region defined on the UV map. 
% % 
% As illustrated in Fig.~\ref{fig:tex-transfer}, we conduct both shape and texture transfer by swapping the corresponding areas on the UV map from the source identity to the target.
% % 
% Visually inspected, the texture-transferred identity inherits the texture characteristics from the source while maintaining the shape layout of the target, and the shape-transferred results maintain the layout of the source without affecting their textures.

% % 
% Furthermore, 
We present the interpolation trajectory of the shape-texture transfer in Fig.~\ref{fig:geo-tex-interp-trajectory}, where both shape and texture are gradually added from the source identity to the target.
The semantically meaningful intermediate results in both shape and texture interpolation validate the effectiveness of our design.

\begin{figure}[h]
  \includegraphics[width=0.5\textwidth]{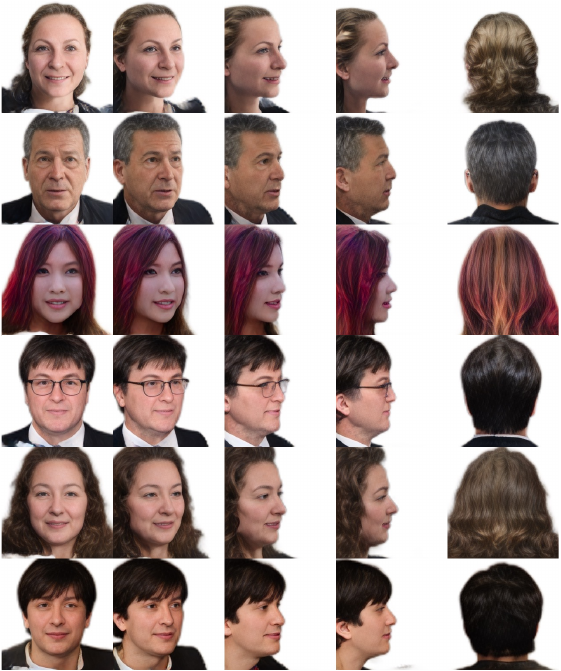} 
  \caption{\textbf{Unconditional Diffusion Sampling.} The compact UV space design allows us to leverage 2D diffusion architectures for 3D aware synthesis.}
\label{fig:diffusion:unconditional}
\vspace{-4mm}
\end{figure}

\heading{Unconditional Generation}
Thanks to the compact UV space design, we can directly leverage powerful 2D diffusion architectures for 3D-aware generation.
Specifically, we train a diffusion model $\model$ over the exported UV maps from the autodecoder $\decoder$ and include the diffusion generation results in Fig.~\ref{fig:diffusion:unconditional}. 
Visually inspected, the diffusion-generated results maintain the same high-fidelity and view-consistent renderings as the reconstruction results with diverse sampling.
Compared with the previous tri-plane-based method~\cite{wang2023rodin,An_2023_CVPR}, \nickname{} maintains high capacity, and flexibility and intrinsically avoids Janus problem.
Besides, diffusion models process better editing ability compared with GAN-based methods.

\subsection{Applications}
% We showcase three applications here to demonstrate the potential of our proposed method.
% 
% First, in~\ref{fig:exp:application:diffusion_inpainting} we show both the geometry- and texture-based 3D-aware inpainting using the pre-trained diffusion model.

\begin{figure}[ht]
  \includegraphics[width=0.5\textwidth]{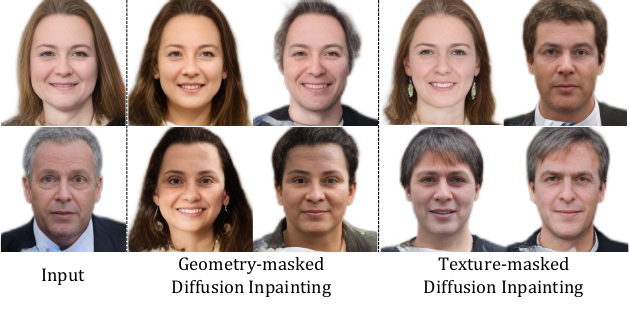} 
\vspace{-4mm}
\caption{\textbf{Diffusion-based Inpainting given Geometry or Texture Mask.} 
We provide either the geometry part (first $9$ channels of the $256\times256\times33$ tensor $\gaussian_0$) or the texture part (last $24$ channels of the $256\times256\times33$ tensor $\gaussian_0$) of the \emph{upper face} as hints, and take the diffusion model $f_\theta$ in-paints the remaining details. 
As in columns $2-3$, all the generated results exhibit an identical upper-face shape, including the layout of the eyes and forehead, but with different textures. Conversely, the diffusion-inpainted results with texture masks (columns $4-5$) showcase an identical upper-face texture, encompassing features such as hair and forehead color, while varying in shape.
  }
\label{fig:exp:application:diffusion_inpainting}
\vspace{-2mm}
\end{figure}

% Both yield holistically reasonable results while keeping the corresponding inputs within the mask unchanged.
\heading{3D Inpainting using Diffusion Model}
% \input{sec/figures/application-diffusion-geo-tex-inpaint}
% \input{sec/figures/application-diffusion-tex-inpaint}
% Our trained diffusion model supports 3D in-painting given geometry and texture mask individually.
% 
First, we showcase both the geometry- and texture-based inpainting in Fig.~\ref{fig:exp:application:diffusion_inpainting}, where unmask the \emph{upper face in the UV space} and let the diffusion model inpaints the remaining areas. 
Both yield holistically reasonable results while keeping the corresponding inputs within the mask unchanged.
% 
% Note that there are still noticeable color changes between the inpainting boundaries, which could be addressed by introducing inpainting-specific training design~\cite{rombach2021highresolution}.

\begin{figure}[ht]
  \includegraphics[width=0.5\textwidth]{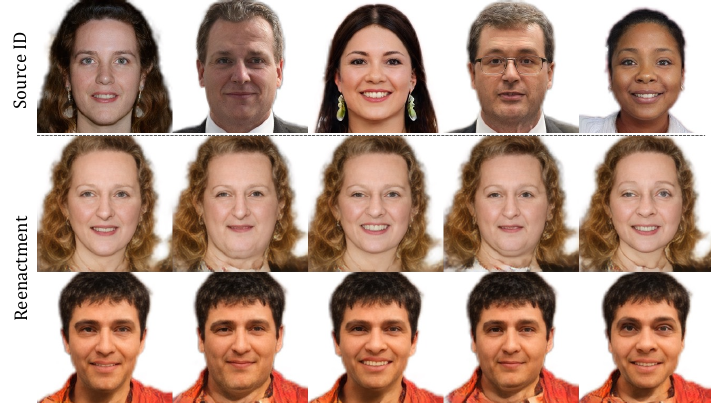} 
  \caption{\textbf{3DMM-based Editing}. We reenact $5$ random sampled source inputs (row-$1$) onto the targets (row-$2,3$) by driving the target Gaussians through the adjustment of the inner 3DMM mesh using the shape and expression codes from the source.
  The reenactment results preserve their original texture while adapting to the shape and expression of the source inputs.
  }
\label{fig:exp:application:3dmm}
\vspace{-4mm}
\end{figure}

\heading{3DMM-based Editing}
\nickname{} marries the best of both the model-based 3DMM and neural representations through the rasterized UV space, and naturally supports 3DMM-based editing, \eg, avatar modeling by changing the shape and expression codes.
We showcase this ability by directly swapping the shape and expression codes of given source identities onto the target instances by driving the learned Gaussians.
As shown in Fig.~\ref{fig:exp:application:3dmm}, the reenactment results maintain their original texture details but accurately follow the shape and expression of the given source input.
This application has the potential to facilitate avatar editing in game engines and media creation. 

\begin{figure}[ht]
  \includegraphics[width=0.5\textwidth]{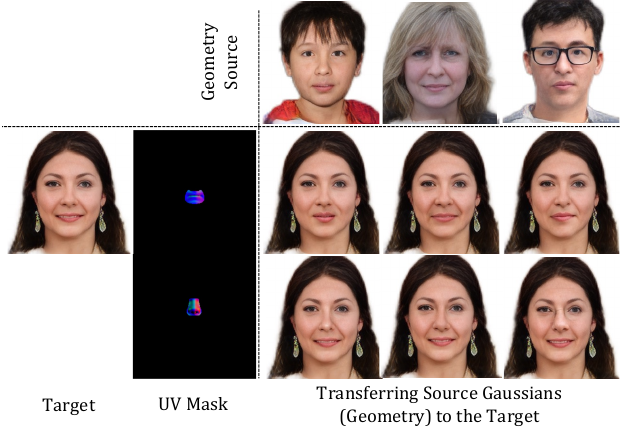} 
\vspace{-4mm}
\caption{\textbf{Regional Editing with UV Mask}. The UV-space design in \nickname{} enables the editing of specific semantic regions defined on the UV map.
  Given diffusion-sampled instances, we transfer the geometry shapes of the "nose" and "mouth" from the source identities to the target while leaving the remaining areas unchanged.
%   
%  Note that all the source and target identities presented here are diffusion-sampled.
  }
\vspace{-4mm}
\label{fig:exp:application:region_transfer}
\end{figure}

\heading{Regional-based Editing}
In addition to global interpolation and transfer capabilities, \nickname{} provides support for region-based editing, allowing modification exclusively within the semantic region defined by the UV mask.
This functionality is illustrated in Fig.~\ref{fig:exp:application:region_transfer}, where we showcase the transfer of corresponding source Gaussians (geometry) to the target, guided by the provided mask.
The transferred results exhibit the same shape as the source within the defined semantic region, while the remaining areas remain unchanged.
Benefiting from the UV space design, regional editing consistently produces semantically consistent results when transferring between mouth/nose regions of varying sizes across different identities.
Furthermore, this demonstration underscores that \nickname{} can surpass 3DMM constraints, enhancing controllability and exhibiting significant potential for avatar personalization within game engines~\cite{metahuman}.

% Please find more results and applications in the supplementary material and demo page.
\subsection{Ablation Study}
\label{subsec:ablation_study}
% We conduct ablation experiments on two design choices within our method: the incorporation of a local tri-plane as the payload and the utilization of a shared convolutional decoder for generating UV feature maps.
We ablate the design choice of adopting local tri-planes as the payload here. Please check the supplementary for more ablations on the utilization of a shared convolutional decoder for generating UV feature maps and the choice of $K$ value in Gaussian blending.

\begin{figure}[ht]
  \includegraphics[width=0.5\textwidth]{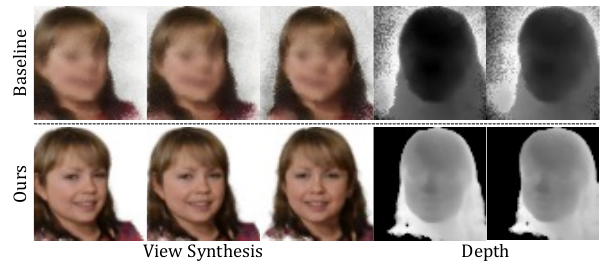} 
\vspace{-6mm}
  \caption{\textbf{Ablation Study on Local Tri-plane.} Using raw feature vectors as the payload lacks the ability to encode spatial information, and our local tri-plane design holds larger capacity and better reconstruction results.}
\label{fig:abla:localtriplane}
\vspace{-4mm}
\end{figure}

\heading{Local Tri-plane}
In our early experiments, we opted for a pure feature vector as the local payload to represent the textures within a local region. 
However, we observed that the reconstruction performance consistently reached limitations, even when overfitting to a single instance. As visualized in Fig.~\ref{fig:abla:localtriplane}, this motivated us to employ a tiny tri-plane as the texture payload. 
For both settings, we utilized $1024$ Gaussians to represent a scene and trained two variations till convergence.
The results indicate that using a pure feature vector as the payload results in blurry view synthesis with a PSNR $21$db and noisy depths.
Conversely, our local tri-plane payload variations exhibit improved fidelity with PSNR $32$db and cleaner surface reconstruction.

\section{Conclusion, Limitation and Future Work}
We have introduced \nickname{}, a new 3D generative framework, and demonstrated its promising results across various scenarios.
We first introduced a novel representation based on 3DMM anchored 3D Gaussians with tri-plane payloads, which allows us to decouple the underlying smooth geometry and deformation from the complex volumetric appearance. 
% 
% This representation decouples the underlying geometry and expression from the complex volumetric appearance and is conveniently stored in the 2D UV space.
% 
Importantly, our representation can be stored in the UV space that is amenable to generative modelling.
We then proposed a method to simultaneously reconstruct and learn a latent space for our 3D representations via multi-view supervision, upon which we train a 2D diffusion model to perform various editing tasks.
We validate our framework on the synthetic dataset based on Panohead \cite{An_2023_CVPR}, which contains diverse, 360-degree view of photo-realistic human heads, though it has very limited variance in expressions.
For future work,
a natural follow up will be to extended our method to full body and introduce text/segmentation control ~\cite{zhang2023adding} on 3D Gaussians.
Moreover, adapting our framework on 3D datasets like ShapeNet~\cite{shapenet2015} and Objaverse~\cite{objaverse} is also meaningful.
Besides, efficient high-res rendering and the support of splatting~\cite{kerbl20233d} is also under-explored.
% Also, we have mostly focused on the generation tasks, we did not pursue efficient design and implementation of our framework and so we are limited to $64 \times 64$ raw render resolution, and consequently we rely on a super-resolution module to produce visually sharp results at the expense of losing some extent of multi-view consistency.
% For the future work we will devote our effort to enable more efficient volume sampling for even faster test time performance.
% Moreover, introducing high-level control mechanisms by introducing text conditions or techniques like ControlNet~\cite{zhang2023adding} on 3D Gaussians is also meaningful.
% Finally, our approach can be extended to full body, where complex motion and interactions of body parts may require finer partitioning of the space, careful 3DMM modelling and compositionality of the representation.
\clearpage
\clearpage
\newpage
{
    \small
    \bibliographystyle{ieeenat_fullname}
    \bibliography{cvpr24.bib}
}


\end{document}